\documentclass[conference]{IEEEtran}
\usepackage{cite}
\usepackage{amsmath,amssymb,amsfonts}
\usepackage{algorithmic}
\usepackage{graphicx}
\usepackage{textcomp}
\usepackage{xcolor}
\usepackage{dblfloatfix}    % To enable figures at the bottom of page
\def\BibTeX{{\rm B\kern-.05em{\sc i\kern-.025em b}\kern-.08em
    T\kern-.1667em\lower.7ex\hbox{E}\kern-.125emX}}
\begin{document}

\title{AI for Investment Fraud Awareness\\}

\author{\IEEEauthorblockN{Prabh Simran Singh Baweja}
\IEEEauthorblockA{\textit{Electrical and Computer Engineering} \\
\textit{CMKL University}\\
Bangkok, Thailand \\
pbaweja@cmkl.ac.th}
\and
\IEEEauthorblockN{Orathai Sangpetch}
\IEEEauthorblockA{\textit{Electrical and Computer Engineering} \\
\textit{CMKL University}\\
Bangkok, Thailand \\
orathai@cmkl.ac.th}
\and
\IEEEauthorblockN{Akkarit Sangpetch}
\IEEEauthorblockA{\textit{Electrical and Computer Engineering} \\
\textit{CMKL University}\\
Bangkok, Thailand \\
akkarit@cmkl.ac.th}
}

\maketitle

\begin{abstract}
In today\textquotesingle s world, with the rise of numerous social platforms, it has become relatively easy for anyone to spread false information and lure people into traps. Fraudulent schemes and traps are growing rapidly in the investment world. Due to this, countries and individuals face huge financial risks. We present an awareness system with the use of machine learning and gamification techniques to educate the people about investment scams and traps. Our system applies machine learning techniques to provide a personalized learning experience to the user. The system chooses distinct game-design elements and scams from the knowledge pool crafted by domain experts for each individual. The objective of the research project is to reduce inequalities in all countries by educating investors via Active Learning. Our goal is to assist the regulators in assuring a conducive environment for a fair, efficient, and inclusive capital market. In the paper, we discuss the impact of the problem, provide implementation details, and showcase the potentiality of the system through preliminary experiments and results.
\end{abstract}


\section{Introduction}
Nowadays, as technology has made investing in the capital market more accessible, an increasing number of people are investing in the market without sufficient knowledge of the stock market.  With the advent of technology, it has become relatively easy to target investors and lure them into fraudulent schemes. To understand the magnitude of the problem, let us look at the recent losses faced by a few countries. In 2018, 197 million pounds were lost in the United Kingdom due to investment scams \cite{b1}. In 2019, investors in Australia lost 63 million dollars from investment scams. These numbers are growing every year due to the extensive use of tactics by scammers through social networks and other online sources. Some of the commonly used tactics for scams are high return, low risk, investing quickly, advertising, fake news. Although there are several articles, books, resources available explaining the common tactics and how to not fall in these tactics, people are still being exploited. Driven by the significance of depth, a question arises: Is there a means for people to be aware of scams and traps without losing money in the market? \\
We present an awareness system to answer this question with the help of Active Learning \cite{b2}. The existing systems like Investmate \cite{b3} make use of Active Learning to equip users with comprehensive knowledge about the stock market, but they do not teach the users about investment fraud. Our awareness system focuses on educating users about scams and traps in the stock market. The awareness system is composed of three components: a personalization engine, a statistical analysis engine, and a learning platform. The personalization engine is the core of the system. It adopts machine learning techniques to predict the investor type. Based on the investor type, game-design elements, scams, and traps are chosen from the knowledge pool. This feedback is transferred to the learning platform to provide a personalized learning experience to the user. \\
In this paper, we focus on building the personalization engine of the awareness system. This is the fundamental step in order to build the entire system. The whole system relies on the accurate prediction of the investor type so that appropriate game-design elements, scams, and traps can be chosen for the user. We developed a prototype and collected data from the prototype. The data is comprised of the digital footprints of users from Thailand, China and India. We designed the experiments to help us decide the top useful features to demonstrate the accuracy of predicting the investor type.
\section{Related Work}
The use of Gamification \cite{b4} to educate people via Active Learning has become increasingly common in this day and age. Gamification is the application of game-design elements and game principles in non-game contexts. There are a few applications launched in the market to educate people about the stock market using Gamification techniques. Some of the top downloaded applications are Investmate \cite{b3}, Stock Market Gamification \cite{b5}. Although these applications follow the strategy of using game-design elements like the trivia questions and rewards to teach the basics of investment, they miss the key part that not all the users are motivated by the same set of game-design elements. Also, the applications do not educate investors about prevalent scams and traps. Our awareness system incorporates a personalized set of game-design elements for each individual and focuses on educating people about the prevalent scams and traps.\\
We adopt machine learning techniques to predict the investor type of users from their digital footprint. Reference \cite{b6} suggests different types of behavior aspects demonstrated by investors. Research \cite{b7} discusses how bias affects the behavior of investors. The strategy used to predict the investor type involves analyzing the behaviors of the users during their interaction with the system. Some of the widely recognized behavior theories are:
\begin{itemize}
\item The investor regret theory is a theory where investors avoid selling a bad investment in order to avoid regret.
\item Mental accounting behavior explains the tendency of humans to compartmentalize different events, and how this impacts investment portfolio.
\item Prospect and Loss Aversion Theory analyzes the change in behavior of investors,  and the degree of emotion they have towards losses and gains.
\end{itemize}
We choose the top metrics from the digital footprint of the individual that reflects the behaviors displayed by them and their knowledge of the stock market. Based on these metrics, we predict the investor type and choose a set of scams and traps from the knowledge pool. \\
We map the set of game-design elements for each investor type from the mapping provided by the domain experts. The approach is to provide the right set of game-design elements to foster the learning process of the user. Research \cite{b8} has shown that the use of varied game-design elements can help trigger different motivational outcomes. In the system, we record all the metrics and extract the important features to analyze the investor type of individuals. 
% Research [Cite research] also shows that Gamification can be a powerful solution to address motivational problems within learning contexts if they are designed with well-established models. Research [Cite research] talks about a few metrics that can help in analyzing the engagement of users in the learning platform. In the system, we record all the metrics and extract the important features to analyze the investor type of individuals.  

\section{Intelligent Gamification Awareness System}
\begin{figure}[htbp]
\centerline{\includegraphics[width=\columnwidth]{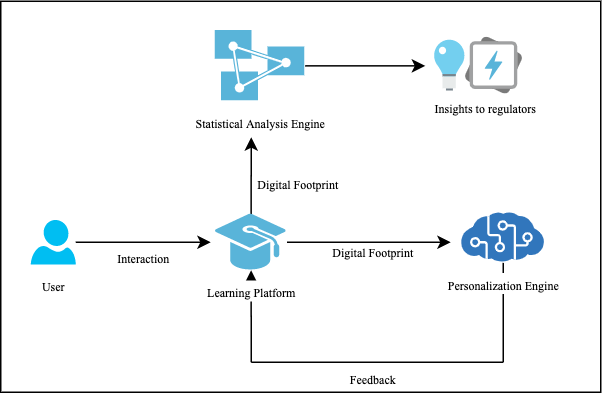}}
\caption{Intelligent Gamification Awareness System Workflow}
\label{fig1}
\end{figure}

The Intelligent Gamification Awareness System comprises of three parts: the learning platform, the personalization engine, and the statistical analysis engine. The user interacts with the learning platform and comprehends the various scams and traps in the stock market. Based on the interaction with the platform, the user’s digital footprint is sent to the personalization engine and the statistical analysis engine. The personalization engine uses various machine learning algorithms to give feedback to the learning platform in order to regularly improve the personalized content provided to the user. The statistical analysis engine provides insights to the regulators to help them adopt better regulation and monitoring policies and strengthen the implementation of such regulations. Fig. \ref{fig1} shows the workflow of the whole system.

\subsection{The Learning Platform}
The idea behind developing the learning platform is to provide a simulation of the stock market and the real world without financial risk to the users. We design the platform to render a personalized learning environment for the user with the integration of various game-design elements and, a set of scams and traps according to the investor type. Each user starts with virtual cash and experience points. The user proceeds to the next level based on their performance on the platform. The following rules \cite{b8} are kept in mind while creating the learning platform:
\begin{itemize}
\item\textbf{Create a user journey}: The platform is not designed in a way that the user walks randomly and does not feel a sense of progression. There are proper onboarding, scaffolding, and pathways to mastery in the platform.
\item \textbf{Balance}: We ensure that there is a balance at all stages in the platform. The difficulty level of the scams and traps depends on the investor type, and they are not too easy or too hard to detect. This is measured by the metrics that record the time spent by the user on the platform for each level.
\item \textbf{Create an experience}: An essential rule of Gamification is to make the platform fun and engaging. We include elements of surprise, winning, problem-solving, exploration in our platform.
\end{itemize}
The platform consists of the following features:
\begin{itemize}
\item \textbf{Personal Portfolio}: All the information related to the assets bought or sold by the investor will be displayed here. The investor gets a sense of accomplishment when they see a profit on their assets. They are motivated to choose a stock wisely if they have a loss in their investments.
\item \textbf{Market Page}: The market page will display information about all of the stocks in the market. There will be a brief overview of the stocks on the market page. Depending on the investor type, the market page might have fake stock companies or fraud companies.
\item \textbf{Stock Details}: The stock details page provides a detailed explanation of every stock, along with the performance of the stock in the past 52 weeks.
\item \textbf{News Page}: The news page will be a collection of news articles based on the stocks available in the market. There is a sentiment (positive, negative, neutral) associated with each news article. The sentiment reflects the reaction of the news article towards the stocks presented in the article. The sentiment is among our list of features collected in the digital footprint of the user. In order to make the platform a simulation of the actual stock market, we add news articles from untrusted and trusted sources. The untrusted sources are news articles with common traps for investors.
\item \textbf{Analytics}: The analytics page gives the user the statistics of their performance in the past. We might provide insights related to other investors in order to motivate the user.
\item \textbf{Chatbot/Mascot}: We introduce a chatbot to the platform to help with the smooth transition of learning about the usage of our platform. The chatbot can be a source of suggestions and insights related to the stocks and news articles. If the user is confused about a stock or a transaction, the chatbot can provide relevant information so that the investor makes an informed decision.
\end{itemize}
The platform records the interactions of the user, and the digital footprint is sent as an input to the personalization engine and the statistical analysis engine. The personalization engine provides continuous feedback to the learning platform. This continuous feedback loop helps in understanding the user adequately, thereby nourishing the learning environment for the user.
\subsection{The Personalization Engine}
The core part of the awareness system is the personalization engine. Figure \ref{fig2} gives an overview of the personalization engine. The user’s digital footprint is provided as the input to the engine. The engine provides continuous feedback to the learning platform. The steps involved in the engine are as follows:
\begin{itemize}
\item \textbf{Feature Extraction}: We use various feature extraction algorithms to extract the top features that provide the most accurate investor type prediction. All the recorded metrics are provided as an input to the feature extraction algorithms. The list of all the recorded metrics in the system are as follows:
\begin{itemize}
    \item Age
    \item Time spent on the fraud stock page
    \item Time spent on the real stock page
    \item Time spent on the fake stock page
    \item Time spent on the market page
    \item Time spent on the portfolio page
    \item Time spent on the news page
    \item Time spent to read positive stock news
    \item Time spent to read neutral stock news
    \item The number of fake stocks bought
    \item The number of fraud stocks bought
    \item The number of real stocks bought
    \item The number of frauds reported
    \item The number of news articles read by the user
    \item The number of asset transactions (buying a stock, selling a stock)
    \item The number of news articles read from untrusted sources
    \item The number of news articles read from trusted sources
\end{itemize}
The top features extracted by the algorithms serve as an input to the investor type prediction algorithms.
\begin{figure}[htbp]
\centerline{\includegraphics[width=5cm, height=10cm]{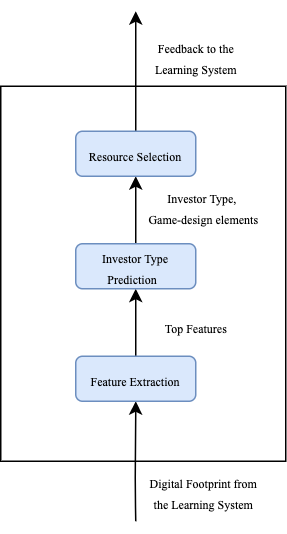}}
\caption{The Personalization Engine}
\label{fig2}
\end{figure}
\item \textbf{Investor Type Prediction}: The input from the feature extraction algorithms is sent to the machine learning algorithms to predict the users into their respective investor types. The underlying mechanism to predict the investor type uses classification algorithms. Based on the investor behavior models \cite{b6}\cite{b7}, we divide the users into five categories. The two broad categories are:
\begin{enumerate}
    \item Novice Investors
    \item Experienced Investors
\end{enumerate}
Experienced investors into further divided into four categories:
\begin{enumerate}
    \item Risk-intolerant traders
    \item Confident traders
    \item Loss-averse young traders
    \item Conservative Long term investors
\end{enumerate}
In order to choose a set of distinct game-design elements for the users, we understand the underlying motivational spectrum of humans \cite{b8}. The spectrum consists of two types of motivations relevant for our awareness platform:
\begin{enumerate}
    \item \textbf{Extrinsic Motivation}: Extrinsic motivation refers to behavior that is driven by external rewards such as money, fame, grades, and praise. This further consists of external regulation, introjection, identification, and integration. The game-design elements that are useful for this category are:
    \begin{itemize}
        \item Badges
        \item Collections
        \item Content unlocking
        \item Leaderboards
        \item Quests
        \item Points
        \item Social Graph
        \item Teams
        \item Virtual goods
        \item Performance-contingent rewards
    \end{itemize}
    \item \textbf{Intrinsic Motivation}: Intrinsic motivation refers to behavior that is driven by internal rewards. The game-design elements suitable for users driven by intrinsic motivation are:
    \begin{itemize}
        \item Quests
        \item Content Unlocking
        \item Performance-contingent rewards
        \item Competence related awards
        \item Unexpected awards
    \end{itemize}
\end{enumerate}
All the above-mentioned game-design elements along with the investor type are passed as an input to the Resource Selection phase. The classification algorithms used for predicting the investor type are Decision Trees \cite{b9}, Gradient Boost Trees (XGBoost) \cite{b10}, and Machine Learning Perceptron \cite{b11}.
\item \textbf{Resource Selection}: The domain experts in the fields of the stock market come up with the knowledge pool of scams and traps prevalent in the stock market. The domain experts also provide a mapping of game-design elements for every investor type. The knowledge pool determines the suitable elements that should be selected according to the predicted investor type of the user.
\end{itemize}
The combination of the game-design elements and, a set of scams and traps is the output of the personalization engine which serves as the feedback to the learning platform.
\subsection{The Statistical Analytics Engine}
The digital footprint of the user is sent as an input to the statistical analysis engine. The engine is responsible for providing insights and feedback to the personalization engine and the regulators of the stock market regularly. We perform descriptive and inferential statistics on the digital footprints of the users and come up with aggregations and interpretations based on the statistics. The aggregations and interpretations assist in formulating insights for the personalization engine. \\
Insights and feedback are also provided to the regulators based on the interpretation of the data. The insights and feedback provided by the statistical analytics engine can be a small step in working towards the United Nation’s Sustainable Goal to reduce inequalities. This can help in the regulation and monitoring of financial markets and institutions.
\section{Prototype}
\begin{figure}[!b]
\centerline{\includegraphics[width=\columnwidth]{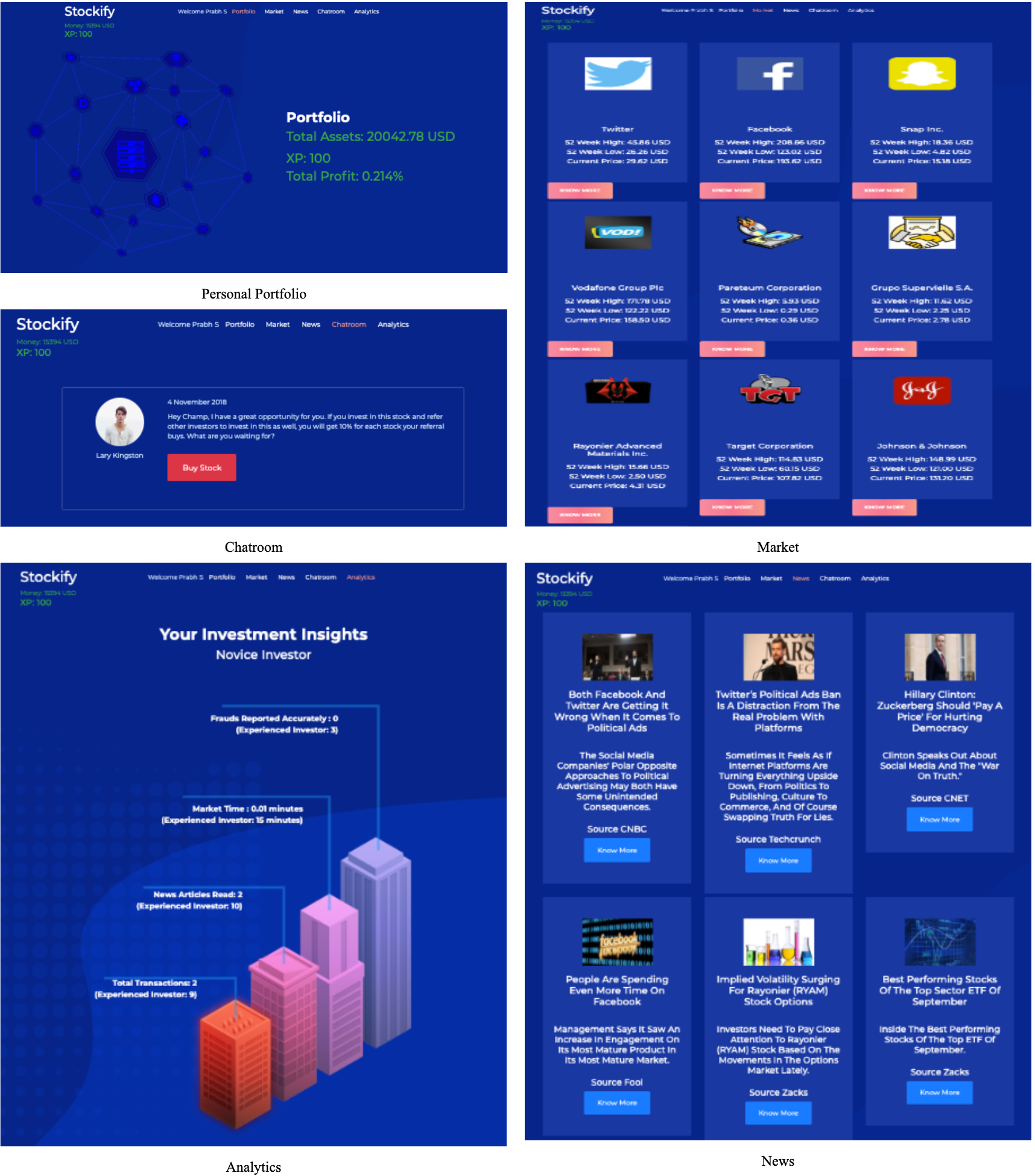}}
\caption{Screenshots of the Prototype}
\label{fig3}
\end{figure}

We have worked closely with the Securities and Exchange Commission, Thailand to develop a prototype to demonstrate the potentiality of our system. With the regular guidance of the officers from the Securities and Exchange Commission, Thailand, we decided to add a couple of traps in our prototype:
\begin{enumerate}
    \item \textbf{Penny Stock Scam}: The scam involves trading stocks of microcap companies. The manipulators and scammers first purchase large quantities of stocks, then drive up the price through false and misleading statements. Then, the manipulators sell their stock at a high price, and all the other stakeholders lose their money.
    \item \textbf{Pyramid Scheme Scam}: A pyramid scheme is a business model that recruits members via a promise of payments or services for enrolling others into the scheme, rather than supplying investments or sale of products.
\end{enumerate}
Fig. \ref{fig3} shows a few screenshots of the prototype. Each user started at the learning platform with virtual money of 20,000 dollars and 100 experience points. Ten stock companies were introduced in the market, among which four of the companies were penny stock fraud companies. News articles related to all the ten companies were published on the news page from trusted and untrusted sources. A chatroom was also added to the prototype to make them aware of the traps used by brokers for the pyramid scheme.
\section{Evaluation}
We conducted a survey to collect information about the background of thirty-three users in order to analyze their interaction with the platform. Apart from the surveys, we interviewed every user to understand their knowledge about the stock market. Ten users in our experiment are from India, twelve are from China, and eleven are from Thailand. Sixteen users had never invested in the stock market, while seventeen users had invested in the stock market.\\
Based on the digital footprint of the users provided by the learning platform, the statistical analytics engine generated insights and shared the following insights with the Securities and Exchange Commission, Thailand:
\begin{enumerate}
    \item Four out of five novice investors were trapped by the penny stock fraud company.
    \item An experienced investor spends two times more time in the market as compared to a novice investor.
\end{enumerate}
To demonstrate the accuracy of the prediction of the investor type by the personalization engine, we perform the following steps:
\begin{itemize}
    \item We run K-Means clustering \cite{b12} on the dataset to determine the optimal number of clusters of investors for the dataset. [Figure 6] shows the results of the elbow method used to analyze the dataset.
    \item Based on the results by K-Means, we divide the users into two broad categories - Novice and Experienced. In the future, when we have more users, we will further divide the experienced investors into subcategories. We divide the thirty-three users dataset in training and test set in a 7:3 ratio. 
    \item We use Principal Component Analysis to extract the top features for determining the investor type based on the digital footprint of the user and the feedback provided by the statistical analysis engine. The top five features are:
        \begin{enumerate}
            \item \textbf{Age}: Age plays a crucial role in the type of stocks an investor buys, and the amount of risk the investor is willing to take.
            \item \textbf{Time spent on the market page}: The time spent on the market page is a clear indicator between a novice investor and an experienced investor.
            \item \textbf{The number of news articles read from untrusted sources}: Novice investors tend to fall into common traps lured by news articles to convince investors to buy penny stocks.
            \item \textbf{The number of fraud stocks bought}: Novice investors tend to fall into penny stock scams easily because of a lack of awareness about the stock market.
            \item \textbf{The number of news articles read from trusted sources}: There is a clear gap between the number of news articles read from trusted sources by experienced investors and novice investors. 
        \end{enumerate}
    \item Based on the top features, we train three different models on the training set and calculate the accuracy of the models on the test set. The personalization engine shows promising results in predicting the investor type. The average accuracy of the different models used in the personalization engine is as follows:
        \begin{enumerate}
            \item Decision Trees: 70\%
            \item Gradient Boost Trees: 80\%
            \item Machine Learning Perceptron: 90\%
        \end{enumerate}
\end{itemize}
\begin{figure}[h!]
\centerline{\includegraphics[width=\columnwidth]{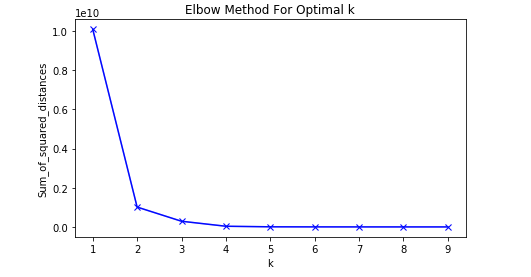}}
\caption{K-Means Elbow Method - Find Optimal Cluster(k)}
\label{fig4}
\end{figure}
The above experiments demonstrate that by utilizing the Machine Learning Perceptron classifier, the personalization engine successfully predicts the investor type with an average accuracy of 90\%. The promising accuracy provides a solid foundation for the further integration of the entire awareness system.
\section{Conclusion}
We introduce an Intelligent Gamification Awareness System to educate people about the common scams and traps in the investment world. The system is designed to provide a personalized learning environment for every user that focuses on educating them about the common scams and traps. Based on the experiments, we demonstrate the accurate prediction of the investor type, which is the core part of the whole system. \\
The Intelligent Gamification Awareness System will be helpful throughout the globe. Given the loss incurred every year on investment scams, our system will play an important role in spreading awareness among investors. The system will perform a significant role in providing insights and feedback to regulators throughout the globe. The system has the potential of assisting regulators towards their goal of providing economic stability to the country.
\section{Future Work}
In this paper, we focus on the working of the core part of our awareness system - the personalization engine. In the future, we will discuss how the feedback from the statistical analysis engine will assist the entire system. We will also demonstrate the effectiveness of our system in raising investors’ awareness of scams.\\
On the other hand, there are numerous potential applications of the Intelligent Gamification Awareness System. The awareness system can be personalized according to the common scams prevalent in the respective nations. This can be a huge step towards reducing inequalities across the globe.\\
Apart from providing awareness on investment scams, the future steps can be to spread awareness about other scams and traps that are getting increasingly common in the global age. The system can be altered to educate people about the widespread fake news, and how to avoid them. The system can also educate people about fake emails and letters that are widespread nowadays. There can be several other use cases to provide awareness of various frauds happening across the globe. Our vision is to help the United Nations towards its goal to adopt better financial policies to achieve greater equality. By 2030, we wish to empower and promote the social, economic, and political inclusion of all people in the world.

% CITATION EXAMPLE
% An excellent style manual for science writers is \cite{b7}.

\section*{Acknowledgment}

 We would like to thank Dr. Chaya Hiruncharoenvate (Officer, Securities and Exchange Commission, Thailand) for providing valuable insights about the investors and common scams that occur in Thailand.

\vspace{12pt}

\end{document}